\documentclass{article} 
\usepackage{iclr2025_conference,times}

\usepackage{amsmath,amsfonts,bm}

\def\eqref#1{equation~\ref{#1}}

\def\1{\bm{1}}

\DeclareMathAlphabet{\mathsfit}{\encodingdefault}{\sfdefault}{m}{sl}
\SetMathAlphabet{\mathsfit}{bold}{\encodingdefault}{\sfdefault}{bx}{n}

\usepackage{mathrsfs} 
\usepackage{longtable}
\usepackage{hyperref}
\usepackage{url}
\usepackage{graphicx}
\usepackage{url}

\title{On Symmetries in Convolutional Weights}

\author{Bilal Alsallakh\\
Voxel AI Labs
\And
Timothy Wroge \\
Voxel AI Labs
\And
Vivek Miglani \\
Meta AI
\And
Narine Kokhlikyan \\
FAIR
}

\iclrfinalcopy
\begin{document}

\maketitle

\begin{abstract}
We explore the symmetry of the mean $k \times k$ weight kernel in each layer of various convolutional neural networks.
Unlike individual neurons, the mean kernels in internal layers tend to be symmetric about their centers instead of favoring specific directions.
We investigate why this symmetry emerges in various datasets and models, and how it is impacted by certain architectural choices.
We show how symmetry correlates with desirable properties such as shift and flip consistency, and might constitute an inherent inductive bias in convolutional neural networks.
\end{abstract}

\section{Introduction}
\label{sec:intro}

\begin{figure*}[!h]
 \centering
 \includegraphics[width=0.8\linewidth]{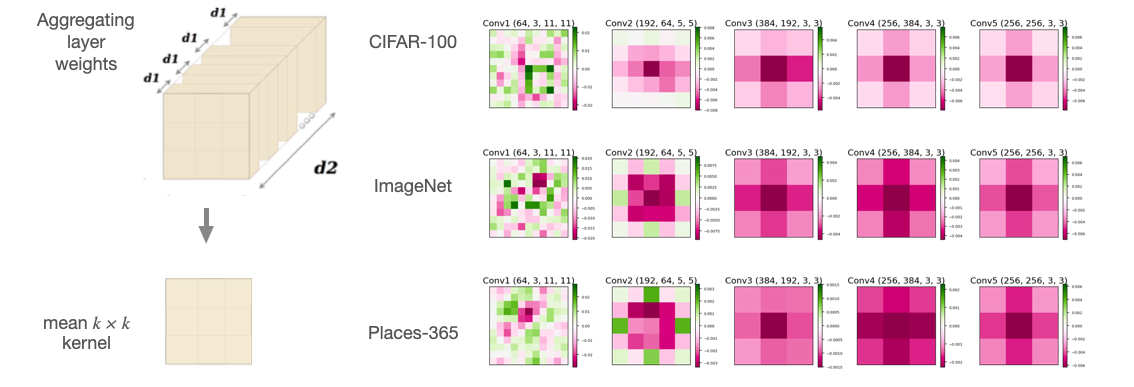}
 \caption {
Left: Illustrating how we compute the mean kernel in a layer.
Right: The mean kernels of AlexNet trained on different datasets.
Title format: (N, C, W, H).
}
 \label{fig:Concept}
\end{figure*}

Various efforts have been made to visualize the weights learned by convolutional neural network, e.g. by treating them as images~\citep{krizhevsky2012imagenet}, via input optimization techniques~\citep{NIPS2016_5d79099f, olah2017feature}, or using matrix factorization~\citep{petrov2021weight, voss2021visualizing}.
In this work we analyze the collective behavior of the weights in each layer, focusing on their symmetry properties in the kernel space.
\cite{mindThePad} noted that the mean kernel in convolutional layers of an ImageNet model tends to be symmetric about its center. We observed this symmetry in a variety of convolutional models trained on various datasets as illustrated in Figure~\ref{fig:Concept}.


To analyze symmetry properties within the weights $W \in R^{N\times C\times H\times W}$ of a convolutional layer, we first compute the mean weight kernel by averaging over the channels $C$ and the neurons $N$.
Appendix~\ref{appendix:graphical_examples} depicts the layer-wise mean kernels of various convolutional models such as different variants of VGGNet~\citep{simonyan2015very}, ResNet~\citep{he2016deep}, and Inception models~\cite{szegedy2016rethinking}.
These models are trained on various image datasets such as CIFAR~\citep{krizhevsky2009learning}, ImageNet~\citep{deng2009imagenet}, and Places365~\citep{zhou2017places}.
It is evident that the mean kernels tend to be symmetric about their centers.

We explore metrics to quantify and analyze the symmetry of layer-wise mean weight kernels in convolutional neural networks.
We investigate the role of the dataset and training data augmentation on the emergence of this symmetry, and how convolutional arithmetic could impact the symmetry at certain layers.
We demonstrate how mitigating unwarranted artificial asymmetry can improve the generalization and robustness of computer-vision models.


\section{Quantifying Symmetry}
We choose to quantify the symmetry of the mean weight kernels based on the transformations of the Dihedral group $D_4$ \cite{dummit2004abstract}. These represent all the possible symmetries of a square (for instance rotation, diagonal splits, etc). We measure the average distance according to the Frobenius norm of the transformed kernel and the original kernel. If a kernel $K$ is perfectly symmetric according to the transformation, then the distance is equal to zero for that particular transformation. By measure the symmetry by averaging over all the possible transformations, as follows: 

\begin{equation}
    S(K) = 1 - \frac{1}{2 \cdot |\mathscr{T}|}\sum_{T \in \mathscr{T}}||T(\hat{K}) - \hat{K}||_F
    \label{eq:symmetry}
\end{equation}
where $\mathscr{T}$ is the set of all transformations or symmetries of the dihedral group $D_4$ excluding the identity element, $||.||_F$ is the Frobenius norm, and $\hat{K} = \frac{K}{|| K||_F}$ is the normalized kernel.
This normalization makes the measure invariant to the magnitude of kernels, which tend to vary across different layers.
More information on the symmetry transformations in $D_4$ is provided in Appendix~\ref{appendix:dihedral}.

\begin{figure*}[!h]
 \centering
 \includegraphics[width=\linewidth]{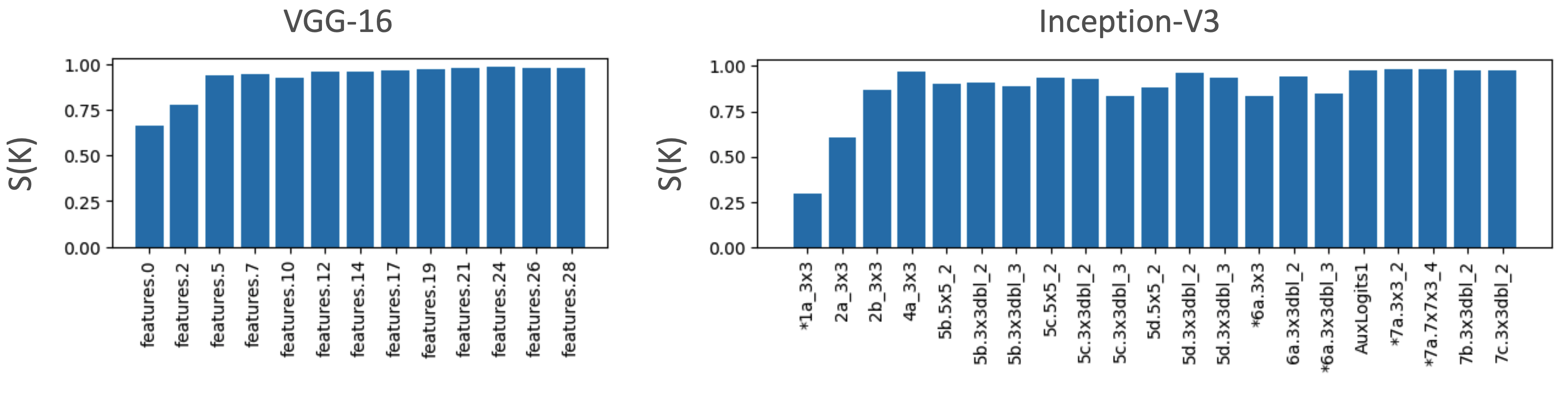}
 \caption {
The symmetry of the mean kernel in each layer of two convolutional architectures, VGG-16 and Inception-V3, according to Eq~\ref{eq:symmetry}. Both models are trained on ImageNet.
}
 \label{fig:symm_profiles}
\end{figure*}

The bar charts in Figure~\ref{fig:symm_profiles} show the value of our symmetry metric for every convolutional layer in two models pre-trained on ImageNet.
It is noticeable in both charts that the  first convolutional layer exhibits lower symmetry than the rest of the layers.
Moreover, the symmetry tends to increase over successive layer, especially for VGG-based models as we demonstrate in Appendix~\ref{appendix:graphical_examples}.

We call the bar chart of per-layer symmetry values in a convolutional model as the symmetry profile of the model.
We next explore why the symmetry emerges, focusing on the role of the dataset and the model architecture.




\section{Role of Dataset and Task}

\begin{figure*}[!t]
\includegraphics[width=\textwidth]{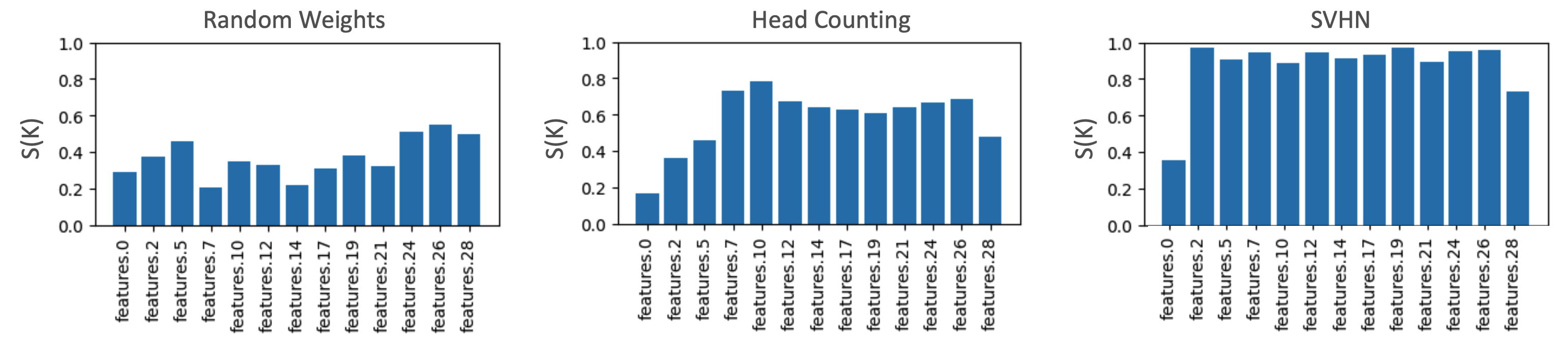}
\label{fig:dataset_impact}
\caption{The similarity profile of a VGG-16 model under various conditions.
Left: The model is initialized with random  weights.
Middle: The model is trained to count heads in an image.
Right: The model is trained on SVHN digit recognition.
}
\end{figure*}

To understand the role of the dataset in the symmetry, we explore the symmetry profiles of various convolutional models trained on various datasets and tasks.
Figure~\ref{fig:dataset_impact} shows examples with a VGG-16 model as implemented in PyTorch's TorchVision library.
As a baseline, we observe the symmetry profile of the model before training, where the waits are initialized using Kaiming's method~\citep{he2015delving}.
This confirms that the symmetry emerges as a result of training the model.

It is noticeable that the symmetry profiles for the same model vary, depending on the dataset and the task.
For example, training the model to count visible heads in an image~\citep{Harinivas2025} does not result in highly-symmetric filters as evident in Figure~\ref{fig:dataset_impact}-middle.
On the other hand, training to classify digits in street signs of house numbers~\cite{netzer2011reading} results in high overall symmetry, with the exception of a few layers (Figure~\ref{fig:dataset_impact}-right).

We observed the symmetry to be consistently high in image-classification models trained on ImageNet, CIFAR, and Places-365.
This suggests that the classification of object and scene images benefits from the emerging symmetry.
These datasets exhibit rotation label symmetry, where a rotated input sample and the original sample have the same class.
The emerging symmetry is in line with the findings by \cite{worrall2017harmonic}, where using rotation equivariant circular harmonics in place of convolution kernels was shown to have higher performance and better generalization in ablation studies. 
For classes of problems where the class label may change upon a transformation, the symmetry of the kernels might not be advantageous, such as alphabet classification.

\section{Role of Model Architecture}

\begin{figure*}[!h]
 \centering
 \includegraphics[width=0.95\linewidth]{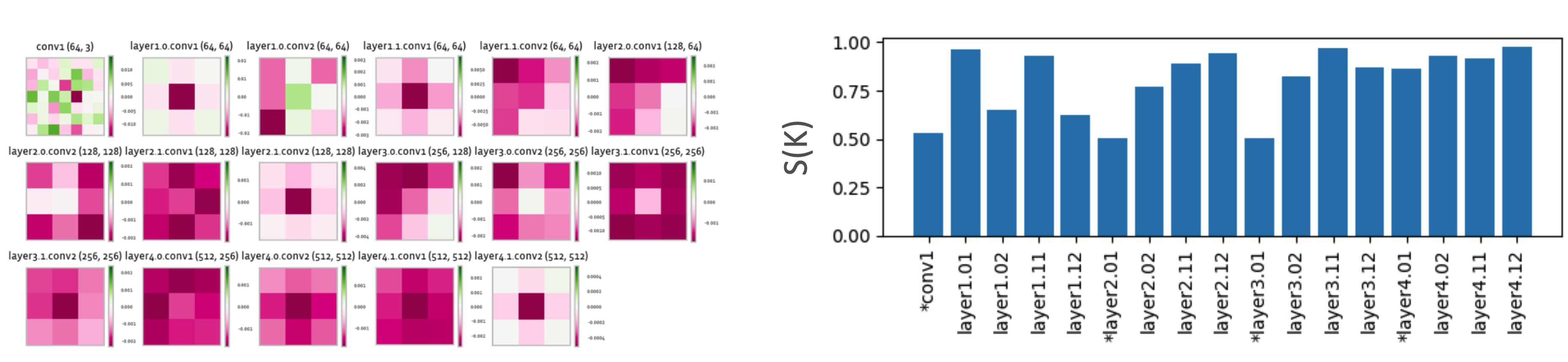}
 \caption {
Left: The mean kernels of a ResNet-18 model, pretrained on ImageNet.
Right: The symmetry of the mean kernels. Notice the significant drops at \texttt{layer2.01} and \texttt{layer3.01}, both of which have a stride of 2 (we indicate that with an asterisk next to the layer name).
}

 \label{fig:resnet_symmetry}
\end{figure*}

Figure~\ref{fig:resnet_symmetry} shows the mean kernels and the symmetry profile of a ResNet-18 model pre-trained on ImageNet, as implemented in PyTorch.
Unlike the previous examples we provided for models pre-trained on ImageNet, the symmetry profile for ResNet-18 exhibits sharp drops in symmetry at certain layers.
At first glance, this seems to contradict our finding about the symmetry profiles of ImageNet classification models, even though these models are trained using the same data augmentations and optimizer.
By exampling the profile closely, we notice that the sharp drops happen mainly at strided-convolution layers, which we indicate with an asterisk next to the layer name.
We report further examples of this observation in a variety of architectures in Appendix~\ref{appendix:asymmetry_examples}.

~\cite{mindThePad} identified uneven application of zero padding as the culprit of artificial asymmetry: The kernels are exposed to zero padding only at the left and top of the input, while the padding at the other sides is ignored due to the arithmetic of strided convolution~\footnote{
The $3\times 3$ kernels in stride-2 convolutional layers need odd numbers as input dimensions in order to consume the padding at all sides, whereas the input image size used for training  is $224 \times 224$.
}.
In contrast, VGG-based models use rigid maxpooling for downsampling instead of learnable strided convolution layers. This avoids artificial asymmetry at the learnable layers where padding is consumed evenly.

There are several approaches to mitigate the aforementioned asymmetry in ResNet models.
For example, using reflection padding or PartialConv~\citep{liu2018partialpadding} eliminates overexposure to zero padding at specific sides.
Likewise, we can change the input size to eliminate uneven application of padding~\citep{mindThePad}.
Furthermore, antialiased CNNs~\citep{zhang2019making} confers significant improvement to the mean-kernel symmetry.
Finally, spatially-balanced pooling (SBPool) was shown to restore mean filter symmetry, even in the presence of uneven padding~\citep{alsallakh2023mind}.
Refer to Appendix~\ref{appendix:asymmetry_mitigation} for mitigation examples on various ResNet models and how we can restore the symmetry of the impacted weight kernels.

\section{Benefits of Symmetry}

ResNet backbones are versatile in CNN-based computer-vision models.
These backbones are prone to artificial asymmetry in the mean kernel, as we illustrated in the previous section.
We demonstrate how mitigating this asymmetry improves model performance and generalization when the task and dataset benefit from the symmetry.
Semantic segmentation is a suited task to showcase such improvements, since it is based on pixel-wise predictions.
These predictions are potentially sensitive to artificial skewness in the weight kernels.

A desirable equivalence property of semantic segmentation is to have horizontal flip consistency: If we flip an image $I$ horizontally, we want the segmentation results $S_{\texttt{flip}(I)}$ to be the flipped version of those for $I$.
We quantify flip consistency by computing the average number of identical pixels between $S_{\texttt{flip}(I)}$  and $\texttt{flip}(S_I)$.
Likewise, if we shift $I$ by a specific amount, we want the segmentation results to be equal to those for $I$ shifted by the same amount after ignoring the vacant shift area.

Table~\ref{tab:segmentation_kpis} shows the consistency results of an FCN segmentation model~\citep{long2015fully} with a ResNet-50 pipeline.
The first row shows results for the baseline pre-trained model as available in PyTorch's TorchVision library.
The symmetry profile for the ResNet-50 backbone it uses exhibits significant drops at strided-convolution layers (see first row in Figure~\ref{fig:ResNet_50_101}).
The model was trained on a subset of the COCO dataset using only the 20 categories that are present in the Pascal VOC dataset. 
We retrain this model under PartialConv~\citep{liu2018partialpadding} which eliminates the use of zero padding and the artificial skewness in the weight kernels as illustrated in the second row of Figure~\ref{fig:ResNet_50_101}.
This results in significant improvement in the model's flip consistency and shift consistency, as evaluated on the validation set. Note that both random horizontal flipping and random crops were part of the training data augmentation used in both the baseline and the retrained models.

\def\arraystretch{1.5}
\begin{table}[h !]
    \centering
    \begin{tabular}{c|c|c|c}
      & mIoU & Flip Consistency (mean) & Shift Consistency  (mean) \\
     \hline
     Baseline FCN & 60.5\% & 91.42\% & 95.42\% \\
     \hline
     FCN with PartialConv & 60.7\% &  93.61\% & 98.06\% \\
    \end{tabular}
    \caption{The mean IoU and consistency performance of two FCN segmentation models. The baseline model uses zero padding, and the second model uses Partial-convolution-based padding.}
    \label{tab:segmentation_kpis}
\end{table}

\section{Discussion and Future Work}
Our work sheds lights into the possible role of mean-kernel symmetry as a useful inductive bias in convolutional neural networks.
This inductive bias improves model robustness to various transforms such as shifts and horizontal flipping, which is a desirable property in various computer-vision tasks.
Further work is needed to understand how the symmetry emerges in the weights, to which extent is it dictated by the dataset and data augmentations, and what other aspects of model architecture and training hyperparameters impact the symmetry.

In our experiments, we noticed that the presence of random horizontal flipping as training data augmentation improves the symmetry.
Moreover, we noticed that overparameterization is necessary for the symmetry to emerge: Drastically reducing the model's capacity seems to make the model compromise on symmetry.
This suggests that symmetry might confer regularization benefits in the presence of overparameterization.
Further work is needed to understand the role of width and depth in fostering symmetry, and how it is impacted by the use of different optimization algorithms.

\bibliography{iclr2025_conference}

\bibliographystyle{iclr2025_conference}


\appendix

\newpage

\section{Examples with Various Models and Datasets}
\label{appendix:graphical_examples}
\begin{figure*}[!h]
 \centering
 \includegraphics[width=\linewidth]{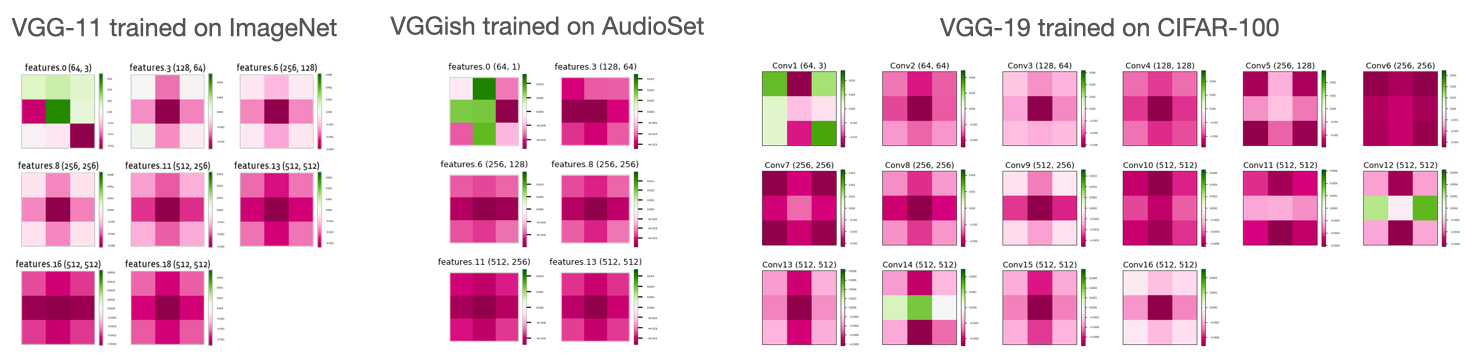}
 \caption {
The mean kernels of three VGG-based models trained on different datasets.
}
 \label{fig:VGGs}
\end{figure*}

\begin{figure*}[!h]
 \centering
 \includegraphics[width=\linewidth]{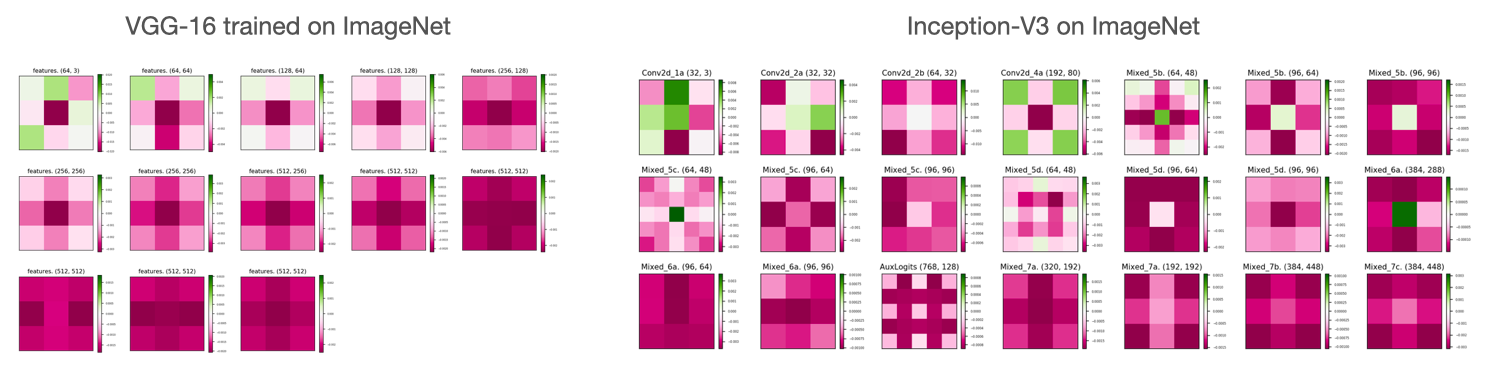}
 \caption {
The mean kernels of two classification models trained on ImageNet.
}
 \label{fig:VGG_Inception}
\end{figure*}

\begin{figure*}[!h]
 \centering
 \includegraphics[width=\linewidth]{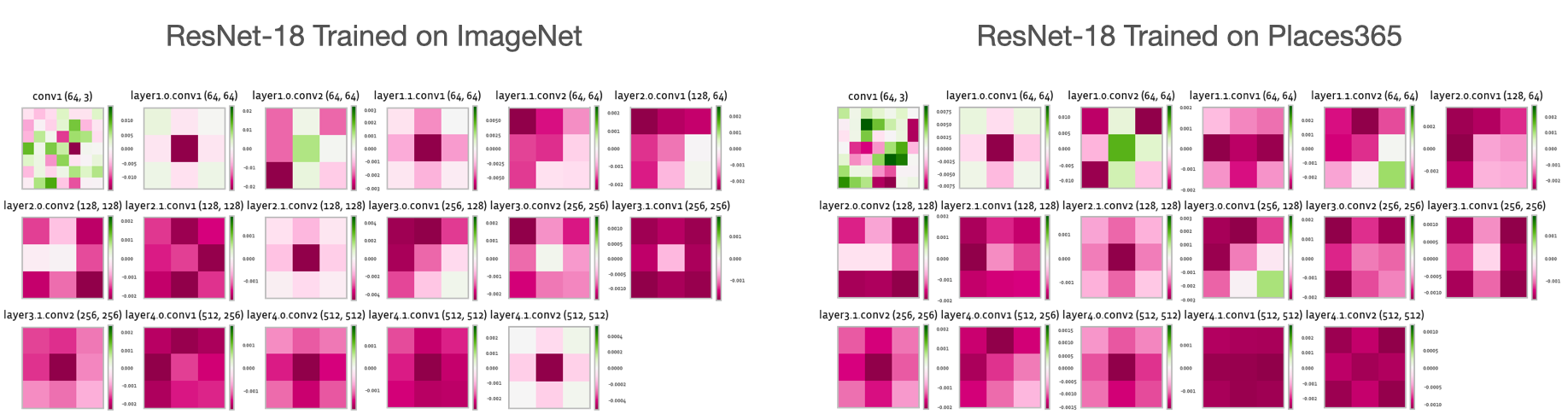}
 \caption {
The mean kernels of ResNet-18 trained on two different datasets.
A few mean kernels exhibit high asymmetry.
 }
 \label{fig:ResNets}
\end{figure*}

Figure~\ref{fig:VGGs} depicts the mean kernels of three publicly-available pretrained models that are based on VGGNet.
The models are trained on various datasets: ImageNet, Audio Set, and CIFAR-100.
Except for the first layer, the mean kernels exhibit high symmetry about their centers.
\newline
The observation holds for various  models such as VGG-16 and Inception-V3 (Figure~\ref{fig:VGG_Inception}).
\newline
Figure~\ref{fig:ResNets} shows further examples of ImageNet classifiers based on ResNet.
While the majority of the internal mean kernels are highly symmetric, a few are remarkably not.

\section{Dihedral Symmetry Group $D_4$}
\label{appendix:dihedral}
The dihedral group D4 represents all possible symmetries of the square. All the transformations together (for instance, rotation, reflection, etc.) form a group. A group is a set of elements that have an identity element, an inverse element, and a binary operation between elements that is also associative. 

D4 retains the order of the vertices of the square so it makes it convenient for matrix operations. For the purposes of quantifying symmetry, the identity does not give any information about the symmetry of the kernel so we omit it. 

Below is a list of the dihedral group's transformations:

\begin{itemize}
    \item $e$: The identity element (no change).
    \item $r$: A $90^\circ$ clockwise rotation.
    \item $r^2$: A $180^\circ$ rotation.
    \item $r^3$: A $270^\circ$ clockwise rotation (or $90^\circ$ counterclockwise).
    \item $s$: A reflection across a vertical axis.
    \item $sr$: A reflection across a diagonal axis from top-left to bottom-right.
    \item $sr^2$: A reflection across a horizontal axis.
    \item $sr^3$: A reflection across a diagonal axis from top-right to bottom-left.
\end{itemize}

Artificial skewness in the learned weights is how a CNN adapts to uneven application of padding and is an artifact of CNN arithmetic.
It does not capture inherent semantic information about the task since it disappears when uneven padding is eliminated. 

\section{Shift Consistency and Weight Symmetry}

\begin{figure*}[!ht]
 \centering
 \includegraphics[width=0.9\linewidth]{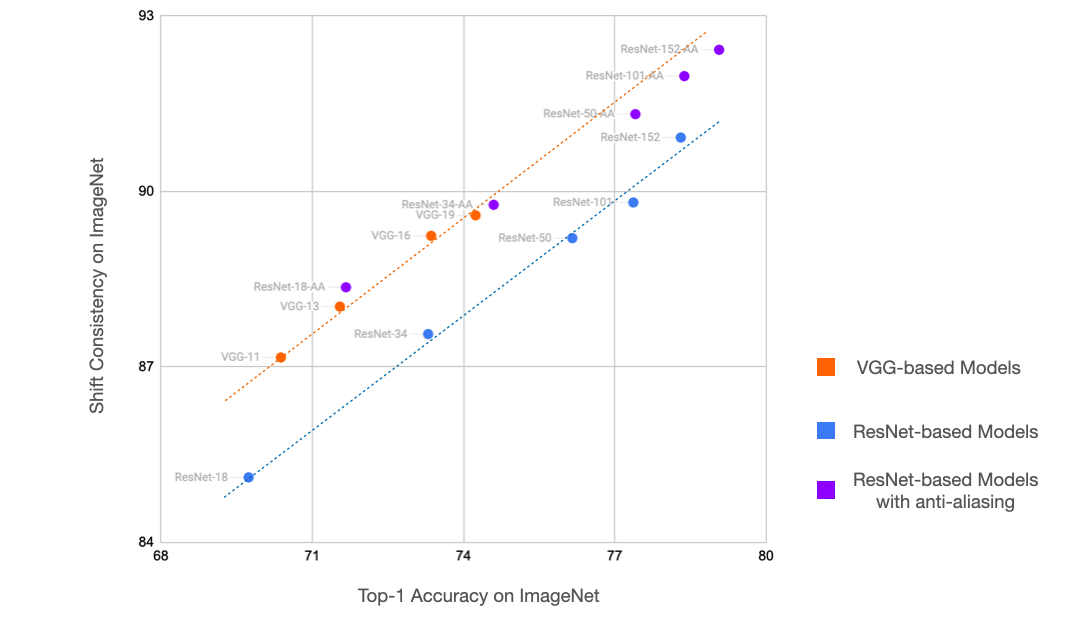}
 \caption {
Analyzing the shift consistency of various VGG-based and ResNet-based models, in relation to model accuracy.
}
 \label{fig:ShiftConsistency}
\end{figure*}

Figure~\ref{fig:ShiftConsistency} plots the shift consistency of various models in relation to their accuracy, as computed by \cite{zhang2019making}.
Expectantly, shift consistency highly correlates with accuracy.
It is noticeable that VGG models exhibit higher shift consistency than standard ResNet models of comparable accuracy.
This is inline with our findings about mean-kernel symmetry in both classes of models, as evident when comparing Figure~\ref{fig:VGGs} and  Figure~\ref{fig:ResNets}.

\begin{figure*}[!ht]
 \centering
 \includegraphics[width=0.95\linewidth]{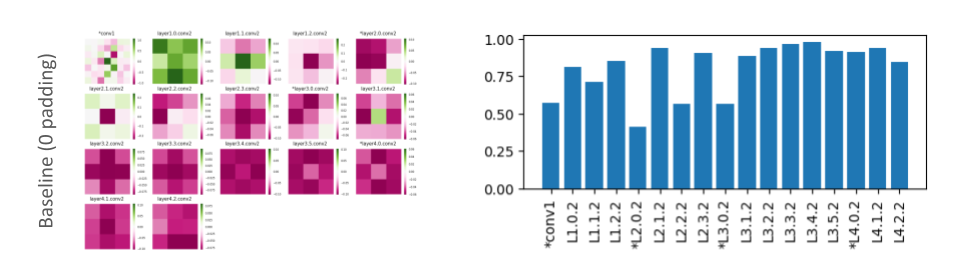}
 \includegraphics[width=0.95\linewidth]{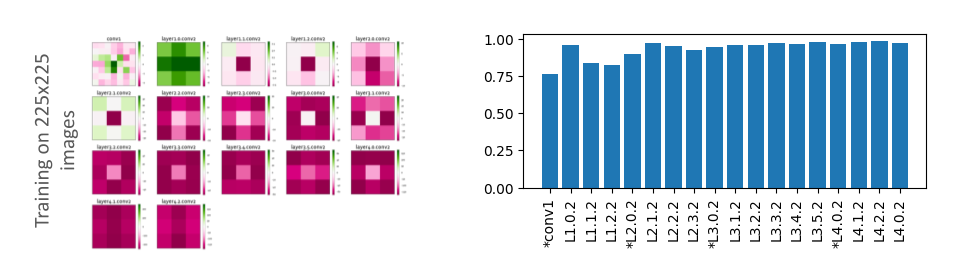}
 \includegraphics[width=0.95\linewidth]{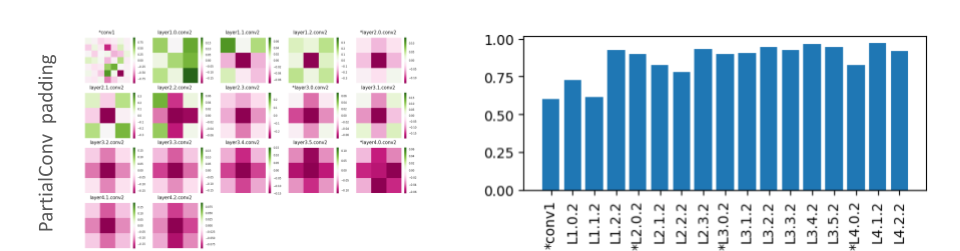}
 \includegraphics[width=0.95\linewidth]{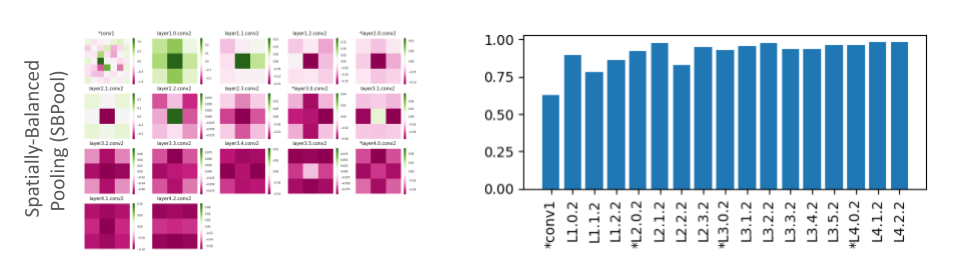}
 \includegraphics[width=0.95\linewidth]{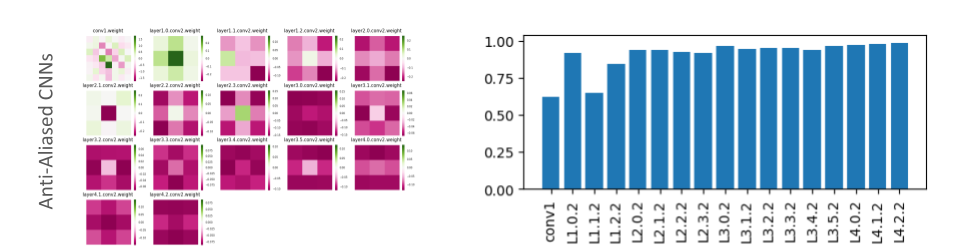}
 \caption {
Four methods for mitigating asymmetry in ResNet-50 trained on IamgeNet.
The first row shows the weight kernel and the corresponding symmetry profile of the baseline model that is trained on $224\times 224$ images under 0-padding. Certain mean kernels at or near downsampling layers (marked with an asterisk) exhibit high asymmetry.
The remaining rows show the same information when training ResNet-50 under various conditions aimed to mitigate the asymmetry: training on $225\times 225$  images~\citep{mindThePad}, training under PartialConv~\citep{liu2018partialpadding}, training under spatially-balanced pooling~\citep{alsallakh2023mind}, and training with anti-aliasing~\citep{zhang2019making}.
}
 \label{fig:ResNet_50_101}
\end{figure*}

The antialiased versions of the ResNet models exhibit improvement both in shift consistency and in accuracy, bridging the gap between the trend lines of standard ResNet and VGG models.
The antialiased models exhibit significantly improve mean-kernel symmetry (Figure~\ref{fig:ResNet_50_101}), which further confirms the role of this symmetry in shift consistency of CNNs.
\clearpage
\section{Asymmetry Examples}

\label{appendix:asymmetry_examples}

\begin{figure*}[!ht]
 \centering
 \includegraphics[width=0.95\linewidth]{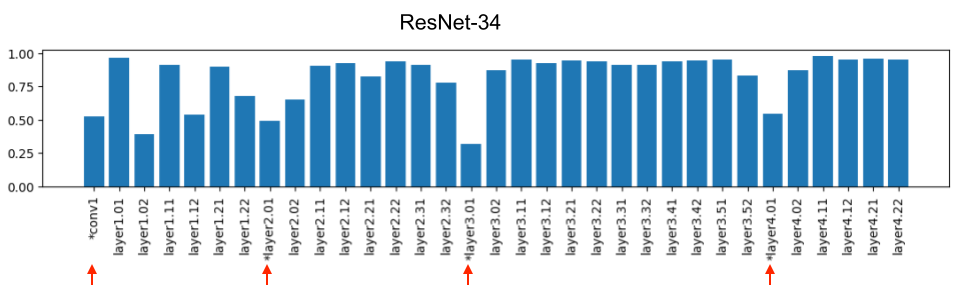}
\vspace{3mm}
\vspace{3mm}
 \includegraphics[width=0.95\linewidth]{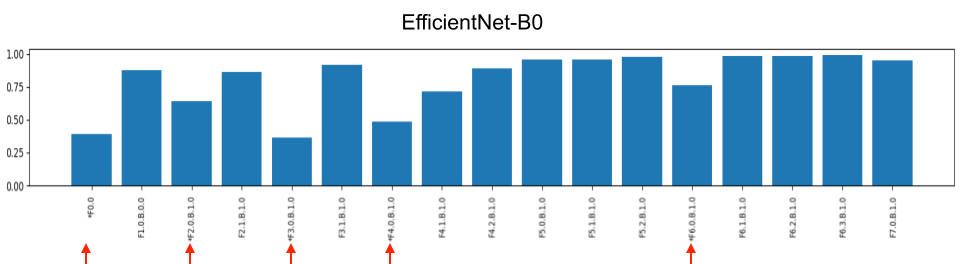}
\vspace{3mm}
\vspace{3mm}
 \includegraphics[width=0.95\linewidth]{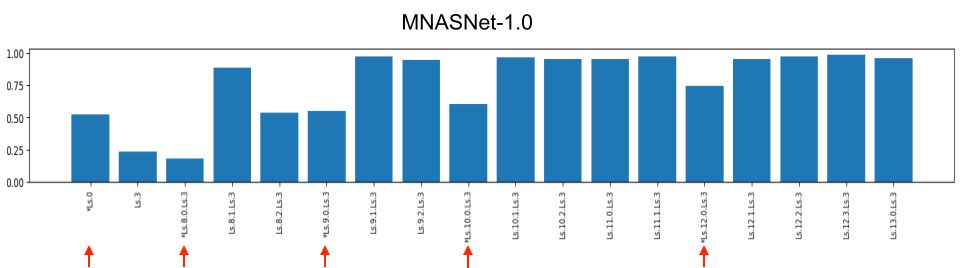}
\vspace{3mm}
\vspace{3mm}
 \includegraphics[width=0.95\linewidth]{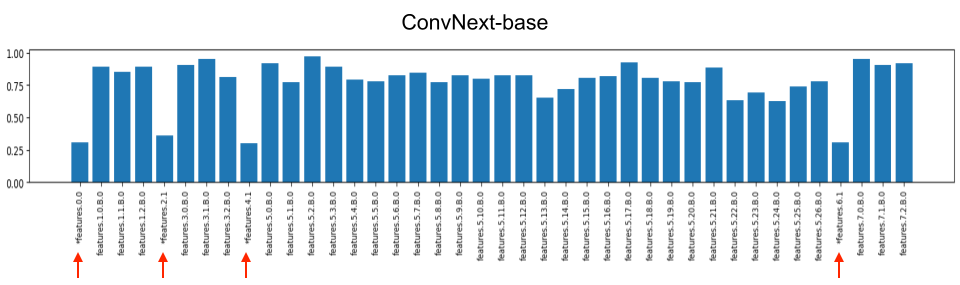}
\vspace{1mm}
 \caption {
Examples of asymmetry at certain layers in various convolutional models, trained on ImageNet, as available in PyTorch's TorchVision library.
The drops in symmetry scores happen mostly at or near downsampling layers, marked with red arrows or with an
asterisk next to the layer's name.
}
 \label{fig:asymmetry_examples}
\end{figure*}

\clearpage
\section{Strategies to Mitigate Asymmetry}

\label{appendix:asymmetry_mitigation}

\end{document}